\documentclass{ws-procs11x85}
\usepackage{ws-procs-thm}           % comment this line when `amsthm / theorem / ntheorem` package is used

\begin{document}

% ========================
% MAIN PAPER CONTENT
% ========================

\title{ColonCrafter: A Depth Estimation Model for Colonoscopy Videos Using Diffusion Priors}

\author{Romain Hardy$^1$$^\dag$, Tyler M. Berzin MD$^2$, Pranav Rajpurkar PhD$^1$}

\address{1. Department of Biomedical Informatics, Harvard Medical School\\
2. Center for Advanced Endoscopy, Beth Israel Deaconess Medical Center}

\begin{abstract}
Three-dimensional (3D) scene understanding in colonoscopy presents significant challenges that necessitate automated methods for accurate depth estimation. However, existing depth estimation models for endoscopy struggle with temporal consistency across video sequences, limiting their applicability for 3D reconstruction. We present ColonCrafter, a diffusion-based depth estimation model that generates temporally consistent depth maps from monocular colonoscopy videos. Our approach learns robust geometric priors from synthetic colonoscopy sequences to generate temporally consistent depth maps. We also introduce a style transfer technique that preserves geometric structure while adapting real clinical videos to match our synthetic training domain. ColonCrafter achieves state-of-the-art zero-shot performance on the C3VD dataset, outperforming both general-purpose and endoscopy-specific approaches. Although full trajectory 3D reconstruction remains a challenge, we demonstrate clinically relevant applications of ColonCrafter, including 3D point cloud generation and surface coverage assessment.

\end{abstract}


% required, do-not-remove
\copyrightinfo{\copyright\ 2024 The Authors. Open Access chapter published by World Scientific Publishing Company and distributed under the terms of the Creative Commons Attribution Non-Commercial (CC BY-NC) 4.0 License.}

\section{Introduction}\label{sec:intro}

Colorectal cancer (CRC) remains a leading cause of cancer-related mortality, with 52,900 projected deaths in 2025 in the United States alone~\cite{siegel2025cancer}. Colonoscopy serves as the gold standard for CRC screening and is designated as a first-tier test by the American College of Gastroenterology~\cite{rex2017colorectal}. During this procedure, gastroenterologists navigate a flexible endoscope through the colon to identify and remove precancerous polyps and other lesions. However, the clinical effectiveness of colonoscopy is constrained by fundamental limitations in human visual perception and spatial reasoning within the complex three-dimensional colonic environment.

These limitations manifest in several critical ways that directly impact patient outcomes. Incomplete examinations occur due to poor visualization behind haustral folds, leading to miss rates of up to 26\% for adenomas~\cite{zhao2019magnitude, thompson2016taller, mathew2021foldit}. Clinicians struggle to relocate previously identified lesions during the same procedure or across multiple sessions, complicating treatment planning and follow-up care~\cite{shandro2020optical}. Perhaps most importantly, accurate measurement of polyp size---a critical factor in determining removal strategy and surveillance intervals---remains challenging using current two-dimensional visualization methods~\cite{o2016accuracy, izzy2015accuracy}. These challenges underscore a fundamental mismatch between the inherently three-dimensional nature of the colon and the two-dimensional visual information available to clinicians.

Bridging this gap between human expertise and the geometric complexity of colonoscopy requires computational tools that can augment clinical decision-making through precise three-dimensional scene understanding. 3D reconstruction of the colon could transform clinical practice by enabling complete surface coverage assessment~\cite{bobrow2023colonoscopy, ma2019real}, accurate lesion localization and size measurement~\cite{soriero2022efficacy}, and robust lesion registration across multiple viewpoints and examination sessions. Such capabilities would complement rather than replace human expertise, providing clinicians with enhanced spatial awareness while preserving their critical role in clinical interpretation and decision-making.

\begin{figure*}[!t]
\centering
\includegraphics[width=\textwidth]{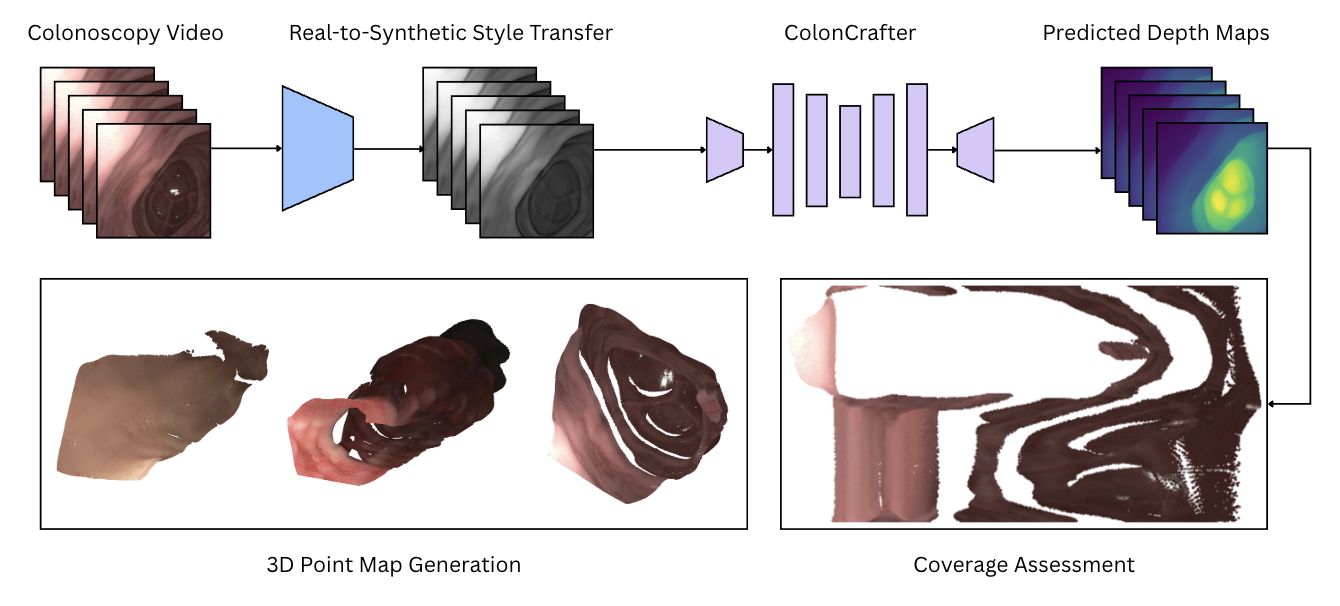}
\caption{Our approach incorporates two key components: (1) ColonCrafter, a diffusion-based depth estimation model trained on large-scale synthetic colonoscopy sequences, and (2) a domain adaptation preprocessing technique that adapts real colonoscopy frames to match the synthetic training domain while preserving geometric structure. ColonCrafter outputs accurate, temporally coherent depth maps suitable for downstream 3D reconstruction and coverage assessment.}
\label{fig:overview}
\end{figure*}

However, achieving reliable 3D reconstruction from colonoscopy videos presents formidable technical challenges that render conventional computer vision approaches ineffective. The colonic environment systematically violates fundamental assumptions underlying traditional Simultaneous Localization and Mapping (SLAM) algorithms~\cite{ma2021rnnslam, golhar2025c3vdv2}. The mucosa lacks distinctive visual features necessary for robust tracking~\cite{ma2021rnnslam, schmidt2024tracking}, exhibits non-Lambertian reflectance with severe specular highlights~\cite{shandro2020optical, pyykola2024non}, and undergoes continuous deformation due to peristalsis and insufflation~\cite{mathew2023self, liu2013robust}. Additionally, rapidly changing illumination from the endoscope's point light source and erratic motion patterns---including rapid rotations, forward-backward movements, and frequent occlusions---further complicate reconstruction efforts~\cite{yao2021motion, golhar2025c3vdv2}.

Recent advances in deep learning have enabled progress toward colonoscopy-specific depth estimation and SLAM systems~\cite{cheng2021depth, mathew2023self, hwang2021unsupervised, ozyoruk2021endoslam}. Self-supervised approaches have shown promise by learning from unlabeled colonoscopy videos~\cite{cheng2021depth, hwang2021unsupervised, lou2024ws}, while others have leveraged synthetic data to overcome the scarcity of ground-truth annotations~\cite{rau2019implicit, mahmood2018deep, itoh2018towards}. Nevertheless, existing methods still struggle with long-term temporal consistency and often fail to generalize across the diverse appearance variations present in clinical data~\cite{ma2021rnnslam, teufel2024oneslam, xu2022deep}. The persistent domain gap between synthetic training data and real colonoscopy imagery represents a critical barrier to clinical translation~\cite{rau2019implicit, he2025synthesized, dinkar2022automatic}, with models trained on synthetic data typically exhibiting poor performance when deployed on real clinical videos~\cite{golhar2025c3vdv2, wang2024structure}.

To address these challenges and enable clinically meaningful AI-assisted colonoscopy, we present ColonCrafter (Figure~\ref{fig:overview}), a diffusion-based depth estimation framework designed to generate temporally consistent depth maps from monocular colonoscopy videos. Our approach addresses the fundamental limitations of existing methods through three key innovations. First, we formulate monocular depth estimation (MDE) as a conditional generation task within a diffusion framework, enabling the model to learn robust priors over the complex appearance and geometry of colonic scenes. Second, we train our model on a large-scale dataset of synthetic colonoscopy sequences derived from computed tomography (CT) scans, providing rich supervision for learning temporally consistent reconstructions. Third, we introduce a style transfer technique that adapts real colonoscopy videos to match the appearance of our synthetic training domain while preserving the geometric structure essential for accurate depth estimation.

Our main contributions advance the integration of AI and clinical expertise in colonoscopy:
\begin{itemize}
\item We present the first diffusion-based depth estimation framework specifically designed for colonoscopy, capable of generating temporally consistent dense depth maps that enable robust 3D scene understanding.
\item We develop a novel style transfer technique that bridges the domain gap between synthetic training data and real colonoscopy videos while preserving geometric cues essential for clinical applications.
\item We demonstrate state-of-the-art zero-shot performance on the C3VD~\cite{bobrow2023colonoscopy} benchmark, showing significant improvements in depth estimation accuracy compared to existing methods.
\end{itemize}

\section{Methods}

We introduce ColonCrafter, a diffusion-based depth estimation model tailored to colonoscopy. Given a monocular colonoscopy video sequence, ColonCrafter estimates temporally coherent dense depth maps, enabling 3D reconstruction of the colon surface. Our approach is based on a video diffusion model trained on a large-scale dataset of synthetic colonoscopy videos derived from CT scans. To overcome the domain gap between our synthetic training data and real-world clinical videos, we further propose a training-free style injection technique that converts any colonoscopy video to the style of our synthetic videos. 

\subsection{Depth Estimation}
\begin{figure*}[!h]
\centering
\includegraphics[width=\textwidth]{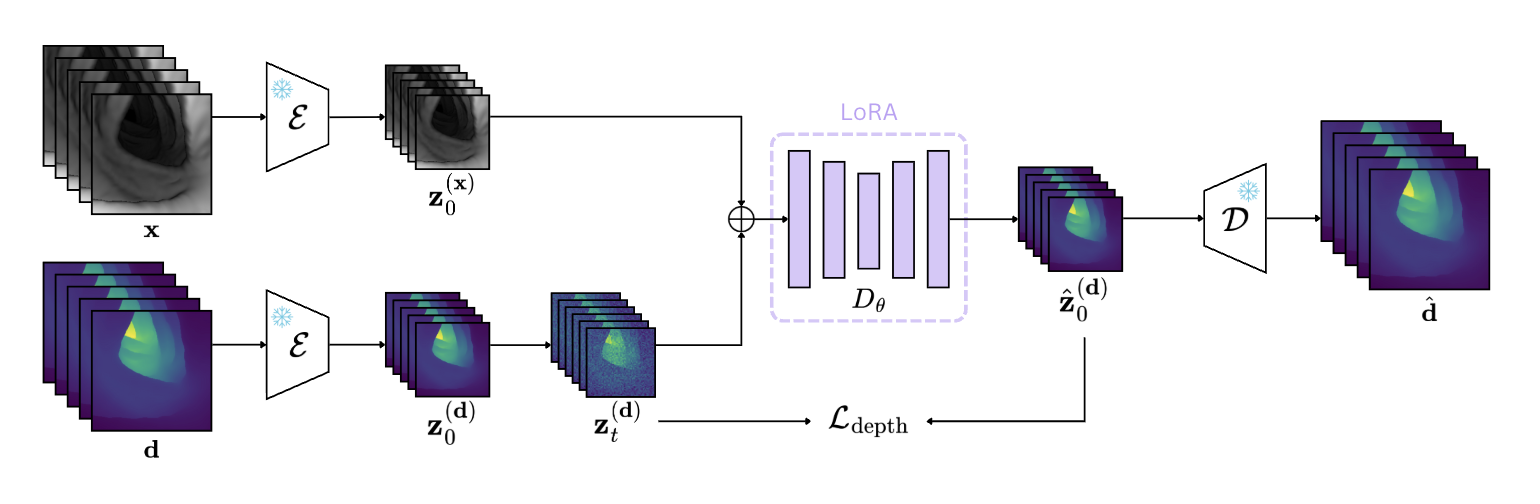}
\caption{Architecture overview of ColonCrafter, a diffusion-based framework for temporally consistent depth estimation in colonoscopy videos. The model employs a conditional diffusion approach where input colonoscopy frames $\mathbf{x}$ are encoded into latent space, and a spatio-temporal U-Net denoiser $D_\theta$ learns to predict clean depth representations from noisy latents conditioned on the input video. Training is performed on synthetic colonoscopy sequences derived from CT scans to learn robust geometric priors.}
\label{fig:coloncrafter}
\end{figure*}

ColonCrafter is a conditional video diffusion model adapted from the DepthCrafter~\cite{hu2025depthcrafter} architecture. Given an input RGB video $\mathbf{x} \in \mathbb{R}^{F \times H \times W \times 3}$, the primary objective of ColonCrafter is to predict a temporally consistent depth sequence $\mathbf{d} \in \mathbb{R}^{F \times H \times W}$. An overview of ColonCrafter is shown in Figure \ref{fig:coloncrafter}.

As with DepthCrafter, we formulate this as a conditional generation task within the EDM \cite{karras2022elucidating} framework. During training, the ground-truth depth volume $\mathbf{d}$ is encoded to a lower-dimensional latent representation $\mathbf{z}^{(\mathbf{d})}_0 = \mathcal{E}(\mathbf{d})$ using the encoder $\mathcal{E}$ of a pre-trained variational autoencoder (VAE). This latent code is then subjected to a forward diffusion process, which progressively adds noise over a continuous time variable $t$:
$\mathbf{z}^{(\mathbf{d})}_t = \mathbf{z}^{(\mathbf{d})}_0 + \sigma_t \epsilon$, where $\epsilon \sim \mathcal{N}(0, I)$ is a Gaussian noise sample and $\sigma_t^2$ is the noise variance at time $t$. Finally, a spatio-temporal U-Net denoiser $D_\theta$ tries to predict the clean latent from the noisy latent: $\hat{\mathbf{z}}^{(\mathbf{d})}_0 = D_\theta(\mathbf{z}^{(\mathbf{d})}_t; \sigma_t; \mathbf{z}^{(\mathbf{x})}_0)$. Here, $\mathbf{z}^{(\mathbf{x})}_0 = \mathcal{E}(\mathbf{x})$ is the encoded latent of the input video $\mathbf{x}$, and serves to condition the denoising process. The denoising objective is given by
\begin{equation}
    \mathcal{L}_{\mathrm{depth}} = \lambda_t\|D_\theta(\mathbf{z}^{(\mathbf{d})}_t; \sigma_t; \mathbf{z}^{(\mathbf{x})}_0) - \mathbf{z}^{(\mathbf{d})}_0\|^2,
    \label{eq:depth_objective}
\end{equation}
where $\lambda_t$ is the loss weight at time $t$. To obtain the predicted depth map $\hat{\mathbf{d}}$, we project $\hat{\mathbf{z}}^{(\mathbf{d})}_0$ back to pixel space using the VAE's frozen decoder $\mathcal{D}$, i.e. $\hat{\mathbf{d}} = \mathcal{D}(\hat{\mathbf{z}}^{(\mathbf{d})}_0)$.

\subsection{Training Details}
\subsubsection{Synthetic Dataset Construction}
\begin{figure*}[!h]
\centering
\includegraphics[width=\textwidth]{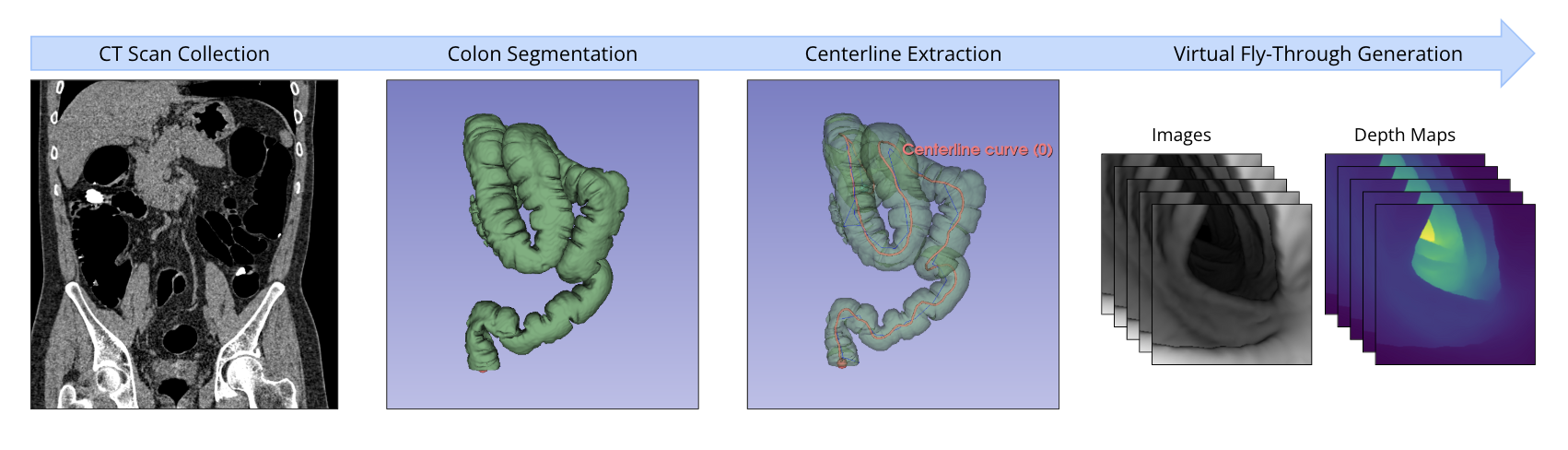}
\caption{Synthetic dataset construction pipeline. We randomly select 5 CT scans from Smith et al.~\cite{smith9data}, segment the colonic volume, extract centerlines, and render virtual fly-throughs to generate synthetic colonoscopy sequences with corresponding ground-truth depth maps.}
\label{fig:dataset}
\end{figure*}

ColonCrafter is trained on 109,329 synthetic colonoscopy images that we generated from 5 CT scans~\cite{smith9data} using standard segmentation and virtual fly-through rendering.~\cite{fedorov20123d, ravi2020pytorch3d} An overview of our dataset construction methodology is shown in Figure \ref{fig:dataset}.

\subsubsection{Model Fine-Tuning}
Rather than training ColonCrafter from scratch, we choose to fine-tune the publicly available checkpoint of DepthCrafter using Low-Rank Adaptation (LoRA) \cite{hu2022lora}. This approach allows us to reuse the high-level features learned during its pre-training phase, while simultaneously reducing the computational cost of adapting the model to the colonoscopy domain. In our implementation, we set the LoRA rank to 16 and target the attention modules of the U-Net denoiser. 

We fine-tune ColonCrafter for 50,000 steps using the AdamW \cite{loshchilov2017decoupled} optimizer with a learning rate of $1.0 \times 10^{-5}$ on a cosine schedule with 1,000 warmup steps. We set the sequence length to 16 and the batch size to 8. To enable ColonCrafter to generalize to complex colonoscopy trajectories, we introduce several novel data augmentation techniques. First, we randomly sample the sequences so that the translations between successive frames are not constant. Second, we randomly flip sections of the video sequences to account for the fact that colonoscopy trajectories are rarely straight paths in practice. Third, we randomly jitter the camera intrinsics to simulate variations in endoscopic equipment. Finally, we apply a random attenuation factor to modify the input video brightness in order to create diverse lighting conditions representative of real colonoscopy procedures.

\subsection{Real-to-Synthetic Style Transfer}\label{style_transfer}
\begin{figure*}[!t]
\centering
\includegraphics[width=0.8\textwidth]{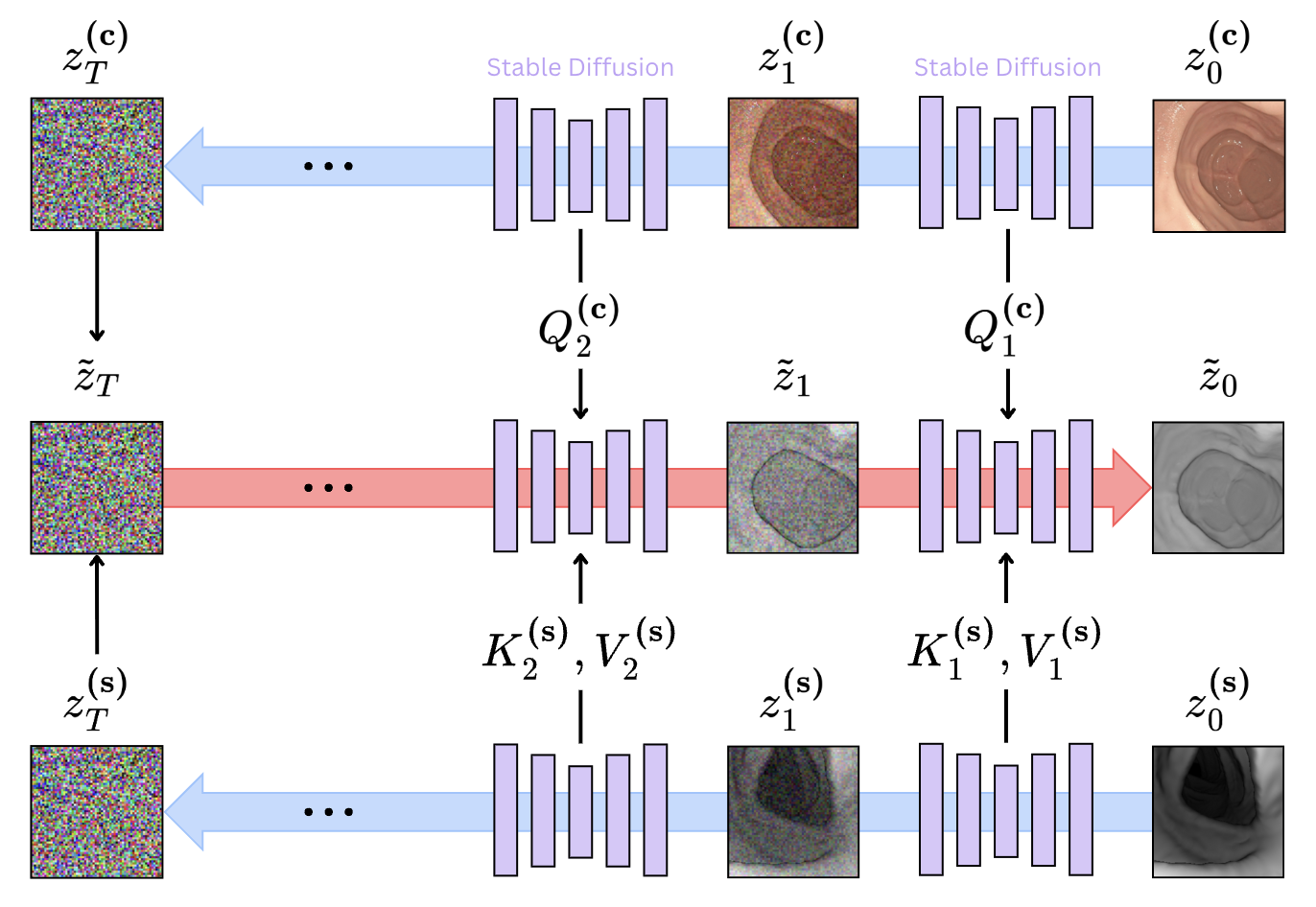}
\caption{Real-to-synthetic style transfer for colonoscopy videos. Given a real colonoscopy image $x^{(\mathbf{c})}$ that we want to convert to the style of a synthetic image $x^{(\mathbf{s})}$, we first invert the latent codes $z_0^{(\mathbf{c})}, z_0^{(\mathbf{s})}$ over $T$ time steps, storing the intermediate key, query, and value vectors at each step before performing reverse diffusion with attention feature substitution.}
\label{fig:style_transfer}
\end{figure*}

Since ColonCrafter is trained on synthetic images with significantly different appearances from real colonoscopy images, we propose a style transfer approach to bridge this gap. Previous works train dedicated neural networks using cycle consistency losses to convert between image domains \cite{mahmood2018unsupervised, mathew2020augmenting, chen2019slam}. However, such models must be trained on large corpora containing sufficiently varied synthetic and real colonoscopy examples, and are prone to destroying structural and lighting cues that are crucial for depth estimation and SLAM. To address these issues, we propose a style transfer approach that enforces content preservation when shifting from the real domain to the synthetic domain, as shown in Figure \ref{fig:style_transfer}. 

Specifically, we cast real-to-synthetic style transfer as a modulation of the denoising process within a pre-trained Stable Diffusion (SD) model, inspired by the work of Chung et al.~\cite{chung2024style} on artistic style transfer. Let $x^{(\mathbf{c})}$ be a colonoscopy video frame that we want to convert to the style of a synthetic video frame $x^{(\mathbf{s})}$. First, we project $(x^{(\mathbf{c})}, x^{(\mathbf{s})})$ to latent codes $(z_0^{(\mathbf{c})}, z_0^{(\mathbf{s})})$ using the VAE encoder of the SD model. Second, we invert the clean latents $(z_0^{(\mathbf{c})}, z_0^{(\mathbf{s})})$ to noisy latents $(z_T^{(\mathbf{c})}, z_T^{(\mathbf{s})})$ over $T$ time steps. At each step $t \in [1, T]$, we store the intermediate queries $Q_{t}^{(\mathbf{c})}$ of the content latent and the keys and values $(K_{t}^{(\mathbf{s})}, V_{t}^{(\mathbf{s})})$ of the style latent. We then carry out the reverse diffusion process with an initial latent input of $\tilde{z}_T = \mathrm{AdaIN}(z_T^{(\mathbf{c})}, z_T^{(\mathbf{s})})$. For every step $t$ of the reverse process and for every self-attention layer in the frozen SD model, we substitute $(K_{t}^{(\mathbf{c})}, V_{t}^{(\mathbf{c})})$ with $(K_{t}^{(\mathbf{s})}, V_{t}^{(\mathbf{s})})$. Since we maintain the original content queries $Q_{t}^{(\mathbf{c})}$ constant, this ensures that we preserve the content cues of $x^{(\mathbf{c})}$ during the denoising process. After completing the reverse diffusion process, we obtain the style-transferred latent code $\tilde{z}^{(\mathbf{c})}$, which we decode into a full-resolution style-transferred video $\tilde{x}^{(\mathbf{c})}$ using the SD model's VAE decoder.

We also introduce three adjustments to facilitate specularity removal and enforce the preservation of depth cues. First, we mask pixels whose intensities exceed $3$ standard deviations from the mean intensity of pixels in a $16 \times 16$ patch. During the reverse diffusion process, the SD model inpaints the masked patches to produce smooth style-transferred surfaces. Second, we apply a local histogram matching approach to align the intensities of the style-transferred videos with those of the original content videos. Third, we truncate the inversion process after $T' = \alpha T$ steps, where $\alpha < 1$. This approach allows us to reduce the amount of noise added during inversion, thus limiting content drift during the denoising process.

\section{Experiments and Results}
\paragraph{Datasets} We evaluate ColonCrafter on 10 colonoscopy sequences from C3VD~\cite{bobrow2023colonoscopy}, a comprehensive dataset featuring realistic colon phantom models with corresponding ground-truth depth annotations. We preprocess all frames following the methodology of Huang et al.~\cite{huang2025advancing}

\paragraph{Baselines and Comparisons} Our evaluation encompasses both general-purpose and endoscopy-specific depth estimation models. For general-purpose baselines, we include the publicly available checkpoints of DepthCrafter~\cite{hu2025depthcrafter} and Depth-Anything~\cite{yang2024depth, yang2024depthv2}. For domain-specific comparisons, we evaluate against three endoscopy-tailored models: EndoDAC~\cite{cui2024endodac}, EndoSfM-Learner~\cite{ozyoruk2021endoslam}, and EndoOmni~\cite{tian2024endoomni}. To ablate our real-to-synthetic style transfer approach, we also evaluate ColonCrafter directly on the colonoscopy sequences without style transfer.

\paragraph{Evaluation Metrics} We assess model performance using standard depth estimation metrics: absolute relative error (AbsRel), squared relative error (SqRel), root mean square error (RMSE), and the $\delta_1$ accuracy measure. Rather than applying per-frame alignment, we compute global scale and shift parameters across each complete video sequence by solving:
\begin{equation}
    \min_{\alpha, \beta \in \mathbb{R}}\sum_{p_i\in\mathcal{P}}[\alpha \hat{d}(p_i) + \beta - d(p_i)]^2,
\end{equation}
where $\hat{d}(p_i)$ and $d(p_i)$ represent predicted and ground-truth depths at pixel $p_i$, respectively, with the summation spanning all pixels $\mathcal{P}$ in the evaluated sequence. To ensure fair comparison, we perform alignment in each model's native training domain---for example, we apply scale-shift alignment in the disparity domain for models trained on disparity data (such as DepthCrafter and ColonCrafter) before converting to the depth domain for final metric computation.

\subsection{Depth Estimation}
\begin{figure*}[!t]
\centering
\includegraphics[width=\textwidth]{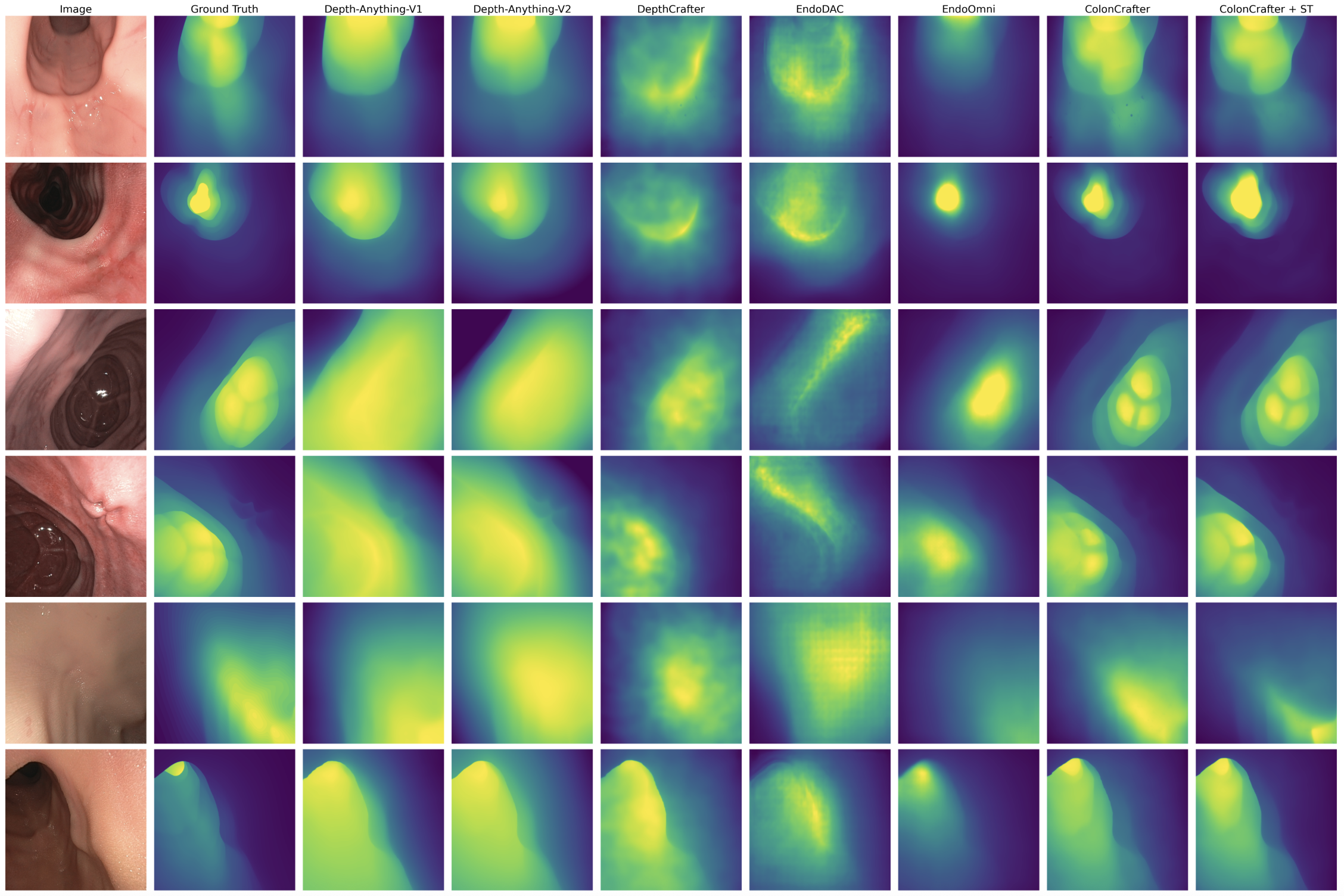}
\caption{Qualitative comparison of depth estimation results on C3VD colonoscopy sequences. Each row shows: (left) original colonoscopy image, (center-left) ground-truth depth map, and depth predictions from various methods including general-purpose models (DepthCrafter, Depth-Anything), endoscopy-specific approaches, and our ColonCrafter with and without style transfer (ST). ColonCrafter + ST produces the most accurate depth maps with sharper boundaries and better preservation of fine geometric details, while effectively handling specular reflections that confuse other methods.}
\label{fig:depth_qualitative}
\end{figure*}

Table~\ref{tab:depth_results} presents the average zero-shot depth estimation performance of ColonCrafter on C3VD. ColonCrafter demonstrates excellent performance, substantially outperforming both general-purpose and endoscopy-specific depth estimation models. Figure~\ref{fig:depth_qualitative} shows qualitative examples of colonoscopy images from C3VD, their ground-truth depth maps, and the depth maps predicted zero-shot by ColonCrafter and other baseline methods.

\begin{table}[t]
\tbl{Zero-shot depth estimation performance on C3VD. We compare ColonCrafter with general-purpose MDE models and endoscopy-specific models. Instead of aligning model predictions with ground-truth labels per frame, we compute a single scale and shift parameter per evaluated trajectory. Parentheses indicate 95\% confidence intervals computed using 1,000-sample bootstrap resampling.}
{\begin{tabular}{@{}lcccc@{}}\toprule
\textbf{Model} & $\delta_1$ $\uparrow$ & \textbf{AbsRel} $\downarrow$ & \textbf{SqRel} $\downarrow$ & \textbf{RMSE (mm)} $\downarrow$\\ \colrule
Depth-Anything-V1~\cite{yang2024depth} & 0.55 (0.45, 0.66) & 0.28 (0.21, 0.34) & 3.58 (1.83, 5.50) & 10.08 (7.08, 13.24) \\
Depth-Anything-V2~\cite{yang2024depthv2} & 0.61 (0.52, 0.71) & 0.24 (0.19, 0.28) & 2.34 (1.55, 3.09) & 8.20 (6.33, 9.87)\\
DepthCrafter~\cite{hu2025depthcrafter} & 0.59 (0.52, 0.65) & 0.22 (0.19, 0.26) & 2.65 (2.07, 3.14) & 10.51 (9.09, 11.70) \\
EndoDAC~\cite{cui2024endodac} & 0.50 (0.42, 0.61) & 0.27 (0.22, 0.33) & 4.45 (2.64, 6.63) & 13.91 (10.45, 17.42) \\
EndoSfM-Learner~\cite{bobrow2023colonoscopy} & 0.56 (0.49, 0.64) & 0.24 (0.21, 0.29) & 3.49 (2.42, 4.68) & 12.34 (9.81, 15.08) \\
EndoOmni$^{\dagger}$~\cite{tian2024endoomni} & 0.77 (0.72, 0.81) & 0.15 (0.14, 0.16) & 1.15 (0.88, 1.43) & 6.91 (5.54, 8.20) \\
ColonCrafter & 0.77 (0.66, 0.83) & 0.16 (0.13, 0.20) & 1.17 (0.81, 1.61) & 6.42 (5.09, 7.62) \\
ColonCrafter + ST & 0.79 (0.70, 0.86) & 0.15 (0.12, 0.18) & 1.09 (0.73, 1.45) & 6.21 (5.01, 7.18) \\ \botrule
\end{tabular}}
\begin{flushleft}
ST: style transfer; $^{\dagger}$: Not strictly zero-shot (partially trained on C3VD).
\end{flushleft}
\label{tab:depth_results}
\end{table}

\paragraph{Comparison with General-Purpose Models} 
Our results reveal a substantial domain gap between open-world and colonoscopy videos. State-of-the-art general-purpose models---including DepthCrafter and Depth-Anything---achieve only moderate performance on colonoscopy data. This performance degradation stems from the unique visual characteristics of endoscopic imagery: constrained lighting conditions, pervasive specular reflections, texture-poor surfaces, and limited field of view present fundamentally different challenges compared to natural outdoor scenes. Our colonoscopy-specific fine-tuning approach effectively overcomes this gap, improving $\delta_1$ accuracy by more than 17\% compared to the base DepthCrafter model. Notably, ColonCrafter outperforms most baseline methods even without style transfer, demonstrating that the model learns robust colonoscopy-specific geometric priors through synthetic data training alone.

\begin{figure}[!h]
\centering
\includegraphics[width=0.5\textwidth]{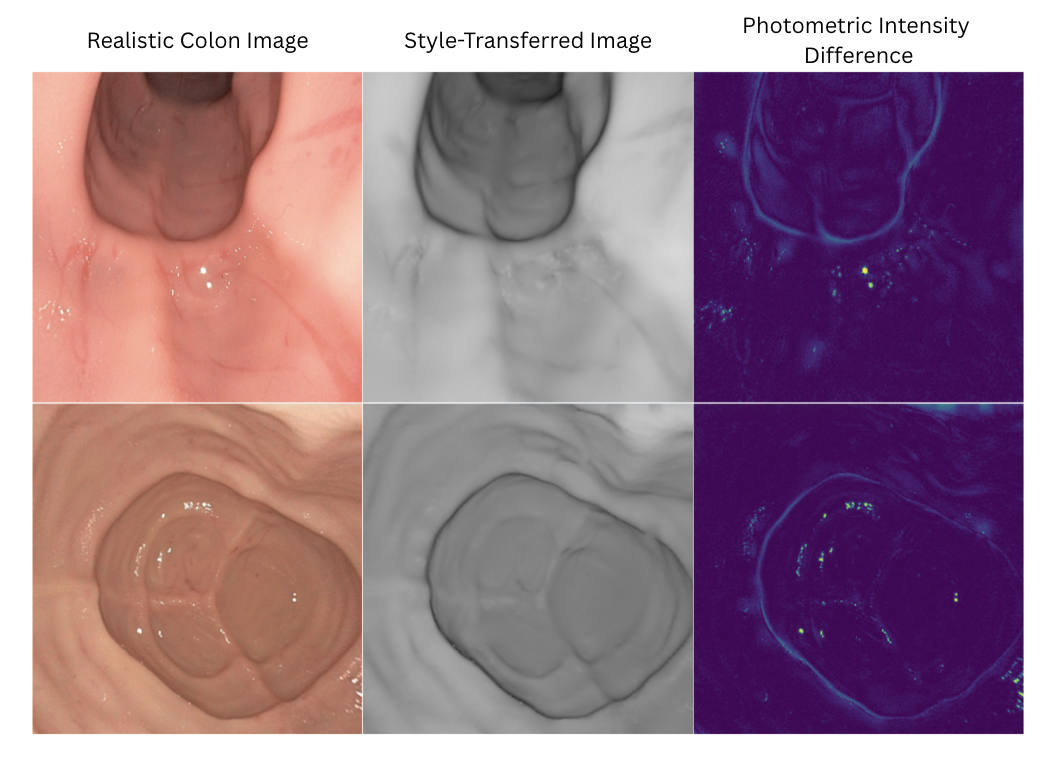}
\caption{Demonstration of real-to-synthetic style transfer on C3VD images. Each triplet shows: (left) original real colonoscopy image with specular highlights and realistic tissue appearance, (middle) style-transferred output that mimics the synthetic data distribution while preserving anatomical structure, and (right) photometric intensity difference map highlighting regions of greatest visual transformation. The style transfer process successfully removes specularities and adjusts texture characteristics to match the training domain, enabling improved depth estimation performance while maintaining essential geometric and structural information.}
\label{fig:style_transfer_qualitative}
\end{figure}

\paragraph{Style Transfer Analysis}
Figure~\ref{fig:style_transfer_qualitative} demonstrates our real-to-synthetic style transfer approach, showing original C3VD images (left), their style-transferred counterparts (middle), and the corresponding photometric intensity difference maps (right). The transformation successfully removes specular highlights and adjusts tissue appearance to match the synthetic training distribution while preserving essential anatomical structure and depth cues. The intensity difference maps reveal that the most significant transformations occur in regions with strong specular reflections and lighting variations, while anatomically critical features such as tissue folds and surface geometry remain largely unchanged. By applying style transfer prior to inference, ColonCrafter operates within the visual domain of its synthetic training data while maintaining the geometric fidelity necessary for accurate depth estimation, resulting in consistent performance improvements observed across all evaluation metrics.

\subsection{Downstream Applications}

\paragraph{Point Cloud Generation} ColonCrafter integrates seamlessly with established SLAM frameworks to produce consistent 3D colon reconstructions. Following the approach of Xu et al.~\cite{xu2025geometrycrafter}, we track feature points across successive frames using a pre-trained SpaTracker~\cite{xiao2024spatialtracker} model. We then estimate camera poses through bundle adjustment by minimizing the reprojection error across all tracked points and frame pairs:
\begin{equation}\label{eq:pose_estimation}
    \min_{W_1,\dots,W_T}\sum_{i,j \in \{1,\ldots, T\}}\|\pi_{K_j}W_jW_i^{-1}\pi_{K_i}^{-1}[p_i, \hat{d}_i] - [p_j, \hat{d}_j]\|^2_2.
\end{equation}
Here, $T$ represents the size of our tracking window, $\pi_{K_i}^{-1}$ denotes the backprojection operator that converts 2D pixel coordinates with depth to 3D world coordinates using camera intrinsics $K_i$, and $W_i$ represents the camera pose (transformation matrix) for frame $i$.

\begin{figure*}[!t]
\centering
\includegraphics[width=0.8\textwidth]{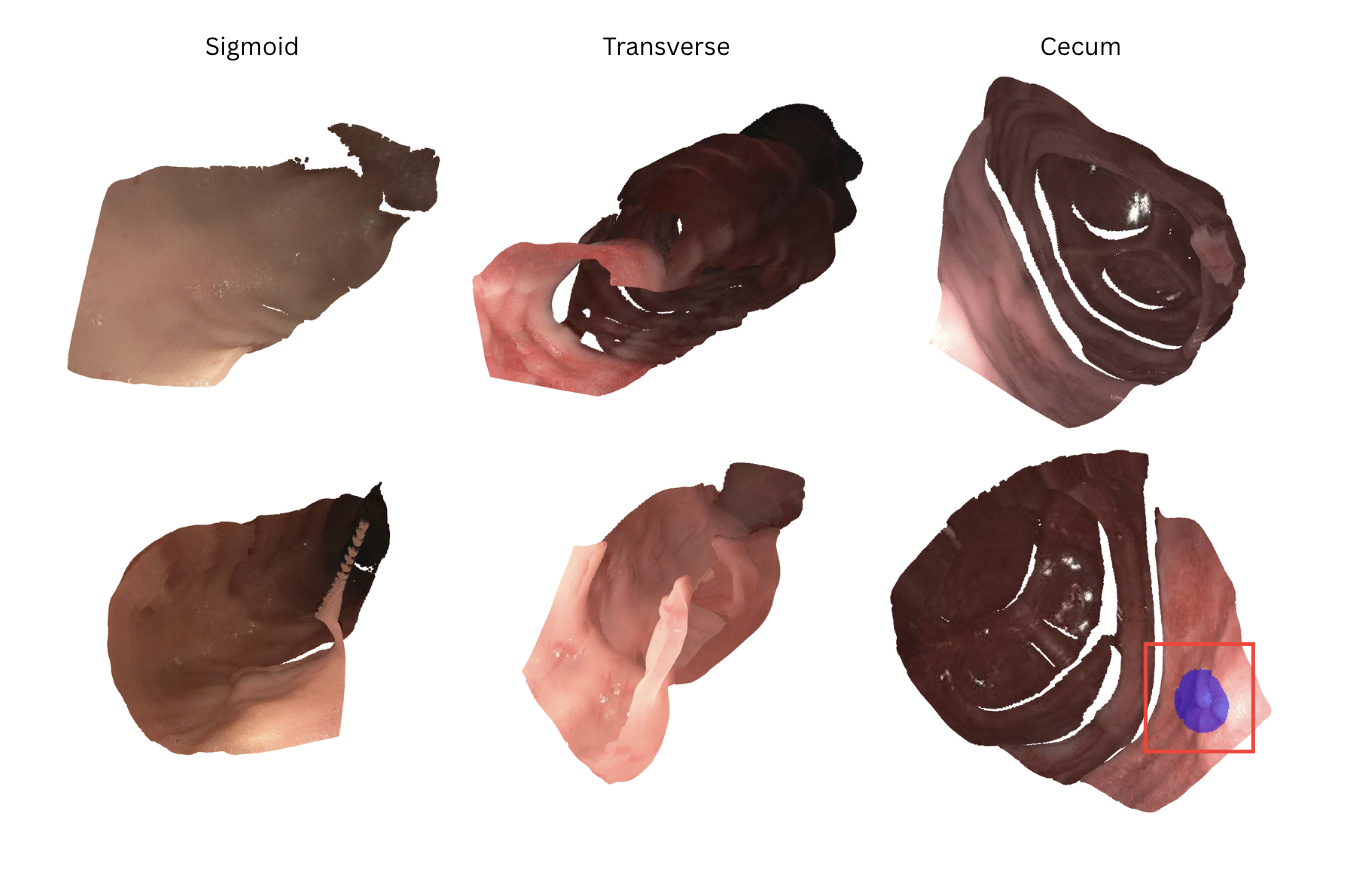}
\caption{3D point cloud reconstructions of colon segments from C3VD sequences using ColonCrafter depth predictions. Each point cloud represents the internal colon structure viewed from different perspectives, with white regions indicating areas not visible during the colonoscopy procedure. Colored annotations (such as the blue protrusion in the bottom right example) highlight anatomical features and potential lesions for clinical analysis. The reconstructions are generated by estimating camera poses through feature tracking and bundle adjustment (Equation~\ref{eq:pose_estimation}), then backprojecting the 2D colonoscopy frames with predicted depth into coherent 3D representations.}
\label{fig:point_clouds}
\end{figure*}

Figure \ref{fig:point_clouds} shows examples of 3D point clouds generated using ColonCrafter for six colonic segments in C3VD. These reconstructions provide detailed structural representations of the colon anatomy, clearly delineating missed surface regions (appearing as white, empty spaces) and lesions. In the bottom right point cloud, we color points belonging to a polyp in blue, demonstrating how point clouds can be used for lesion registration. This capability has significant clinical potential for improving the accuracy of longitudinal lesion assessment by facilitating precise lesion tracking and size measurement in follow-up colonoscopies.

\subsection{Missing Surface Estimation}
\begin{figure*}[!h]
\centering
\includegraphics[width=0.8\textwidth]{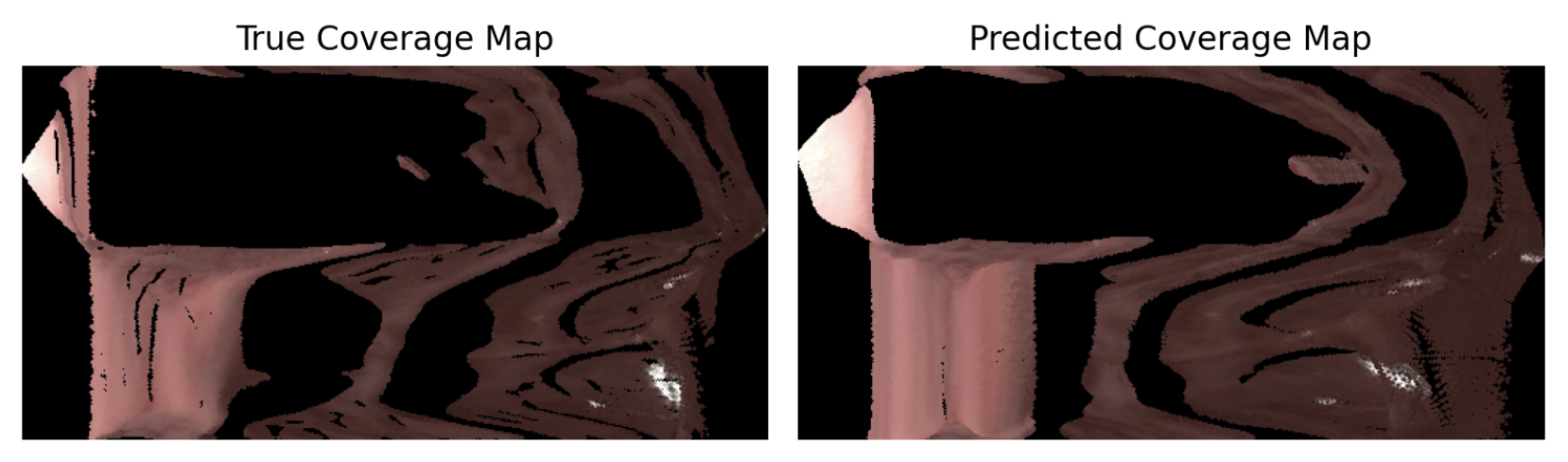}
\caption{Colon surface coverage analysis. Coverage maps show the surveyed areas of colon surface ``unrolled" into 2D representations, where the horizontal axis represents distance along the estimated centerline and the vertical axis represents circumferential angle. Black regions indicate areas unseen during the colonoscopy sequence, while colored regions represent seen surfaces. Comparison between ground-truth depth-based coverage (left) and ColonCrafter prediction-based coverage (right) demonstrates the model's accuracy in identifying unseen areas.}
\label{fig:coverage}
\end{figure*}

Using the 3D point clouds generated by ColonCrafter, we can quantitatively assess unseen areas of the colon in an image sequence. For a given point cloud, we estimate its centerline using Principal Component Analysis, then ``unroll" it into a 2D coverage map where the horizontal axis represents the distance along the centerline and the vertical axis represents the circumferential angle around it. From this representation, we calculate the coverage ratio as the fraction of surveyed pixels after applying morphological cleaning operations to remove noise and artifacts. Figure~\ref{fig:coverage} shows coverage maps computed using ground-truth depths (left) and ColonCrafter predictions (right), demonstrating the model's ability to accurately identify missed surfaces.

\section{Related Work}\label{sec:related_work}

\subsection{Monocular Depth Estimation}

Monocular depth estimation addresses the inherently ill-posed problem of predicting dense depth maps from single RGB images, where a single image can correspond to infinite 3D scenes. Traditional geometric approaches to MDE often produce sparse outputs and are sensitive to lighting conditions, leading to the development of learning-based methods. Recent advances include open-world foundation MDE models such as Depth-Anything\cite{yang2024depth, yang2024depthv2} and DepthCrafter\cite{hu2025depthcrafter}, and endoscopy-specific approaches like EndoDAC\cite{cui2024endodac} and EndoOmni\cite{tian2024endoomni}.

\subsection{Image Style Transfer}

The persistent domain gap between synthetic training data and real colonoscopy imagery has motivated the adoption of style transfer techniques for domain adaptation in gastrointestinal endoscopy~\cite{rau2019implicit, dinkar2022automatic}. Recent diffusion-based style transfer methods offer compelling training-free alternatives that better preserve content structure while enabling domain adaptation~\cite{zhang2023inversion, chung2024style, wang2023stylediffusion}. Chung et al.~\cite{chung2024style} proposed manipulating self-attention layers in pre-trained diffusion models through key-value substitution while preserving query information, explicitly maintaining source image content during stylization. This paradigm is particularly appealing for colonoscopy applications where preserving geometric and depth cues is essential for accurate 3D reconstruction. However, despite their theoretical promise, these techniques remain largely unexplored in endoscopic applications.

\section{Discussion}
\paragraph{Clinical Significance}
Our work addresses a fundamental challenge in colonoscopy: reconciling the three-dimensional nature of colonic anatomy and the two-dimensional visual information available to clinicians. ColonCrafter's ability to generate temporally consistent depth maps represents a significant step toward enhancing clinical decision-making through computational augmentation of human expertise. The demonstrated improvements in depth estimation accuracy translate to meaningful clinical benefits for lesion localization, size measurement, and coverage assessment during routine colonoscopic examinations.

\paragraph{Technical Contributions}
The integration of diffusion models with colonoscopy-specific constraints represents a meaningful advance in generative approaches to medical depth estimation, enabling temporally consistent depth estimation that outperforms both general-purpose and endoscopy-specific methods. Our style transfer approach demonstrates a novel paradigm for overcoming domain gaps in medical imaging without using additional data. 

\paragraph{Limitations and Future Work}
Several limitations present opportunities for future development. First, our evaluation on C3VD phantom data provides initial validation, though clinical utility requires validation on real patient procedures with diverse anatomical variations and pathological conditions. Second, ColonCrafter is not designed for full-length colonoscopy procedures, as it operates optimally on shorter video segments due to computational and memory constraints. Finally, we plan to extend ColonCrafter to a semi-supervised framework that trains jointly on real and synthetic colonoscopy sequences. This approach would reduce dependency on synthetic data and enable better generalization to diverse clinical scenarios.

\section{Conclusion}
We present ColonCrafter, a novel diffusion-based depth estimation model specifically designed for colonoscopy videos that achieves state-of-the-art zero-shot performance on the C3VD dataset. Our approach successfully addresses the challenge of temporal consistency in medical depth estimation by combining video diffusion modeling with a style transfer technique that overcomes the domain gap between synthetic training data and real colonoscopy images. ColonCrafter demonstrates substantial improvements over both general-purpose and endoscopy-specific baselines, with particularly strong performance in handling challenging visual characteristics such as specular reflections and complex tissue geometry. While the current framework is optimized for shorter video segments with smooth camera trajectories, it establishes a strong foundation for future developments in computational colonoscopy assistance and demonstrates that synthetic training data can be effectively leveraged for real-world medical applications through appropriate domain adaptation techniques, enabling practical applications such as 3D reconstruction, lesion localization, and surface coverage analysis.

\section*{Acknowledgments}
This is a preprint of a paper accepted for publication in \textit{Proceedings of the Pacific Symposium on Biocomputing (PSB 2026)}, published by World Scientific. © 2026 World Scientific Publishing. Once published, the definitive version of the paper will be available via its DOI.

\bibliographystyle{ws-procs11x85}
\bibliography{bibliography}

\end{document}